\documentclass[manuscript]{acmart}
\usepackage{amsmath}   
\usepackage{mathtools}
\usepackage{soul}      
\usepackage{hyperref}
\usepackage{threeparttable}

\AtBeginDocument{%
  }

\begin{document}

\sloppy

\title{Reweighting is not Harmless: Understanding the Role of Dataset Characteristics}


\author{Amalia Ionescu}
\affiliation{%
 \institution{Cedars-Sinai Medical Center}
 \city{Los Angeles}
 \state{California}
 \country{USA}}
\email{amalia.ionescu@csmc.edu}

\author{Jose G. Hernandez}
\affiliation{%
 \institution{Cedars-Sinai Medical Center}
  \city{Los Angeles}
 \state{California}
 \country{USA}}
\email{jose.hernandez8@csmc.edu}

\author{Emily F. Wong}
\email{emily.wong@cshs.org}
\affiliation{%
 \institution{Cedars-Sinai Medical Center}
 \city{Los Angeles}
 \state{California}
 \country{USA}}

\author{Anil K. Saini}
\affiliation{%
 \institution{Cedars-Sinai Medical Center}
  \city{Los Angeles}
 \state{California}
 \country{USA}}
\email{anil.saini@cshs.org}

\author{Jason H. Moore}
\affiliation{%
 \institution{Cedars-Sinai Medical Center}
  \city{Los Angeles}
 \state{California}
 \country{USA}}
\email{jason.moore@csmc.edu}

\author{Tiffani J. Bright}
\affiliation{%
 \institution{Cedars-Sinai Medical Center}
  \city{Los Angeles}
 \state{California}
 \country{USA}}
\email{tiffani.bright@csmc.edu}



\begin{abstract}
Reweighting is a widely used pre-processing method for promoting discrimination-free model behavior by assigning different weights to data points during model training. While it is often credited with improving fairness, particularly by mitigating systematic underestimation for underrepresented groups, it can also introduce unintended harms under certain data conditions. Using controlled simulations that vary group size and baseline event prevalence, we show how reweighting can yield misleading improvements in fairness by reducing underestimation for one group while introducing it for another. Across 3,000 simulated datasets with event prevalence of 7--14\%, reweighting improved parity in positive classification and false negative rates (FNR) between groups, but did so partly by decreasing positive classifications and increasing FNR for the group with higher baseline prevalence. These effects were most pronounced when that group was a minority: its FNR rose by 6 percentage points, approximately 82 additional missed cases per 10,000 group members, and roughly three-quarters of the improvement in FNR parity came from this increase rather than from gains for the other group. The pattern held across six classifiers, from logistic regression to neural networks, while overall accuracy was essentially unchanged, leaving the shift invisible in aggregate performance metrics. These findings mirror patterns reported in real-world applications and underscore the critical role of data characteristics in shaping the impact of this fairness intervention. While reweighting may create desirable outcomes in some applications, practitioners should critically evaluate it relative to their problem area and goals. Specifically, when the goal is to reduce underestimation in one group without introducing underestimation in another, reweighting may not be the appropriate approach.
\end{abstract}

\ccsdesc[500]{Computing methodologies~Simulation evaluation}
\ccsdesc[500]{Computing methodologies~Machine learning approaches}
\ccsdesc[500]{Computing methodologies~Modeling methodologies}
\ccsdesc[300]{Applied computing~Health informatics}
\keywords{Fairness, Reweighting, Bias Mitigation}


\maketitle

\section{Introduction}
Artificial intelligence (AI) and machine learning (ML) tools are increasingly shaping many aspects of our daily lives, from determining credit lending \cite{bhutta2021} to recommending patients in need of extra care \cite{obermeyer2019}. As these tools become more embedded in high-stakes decision-making, ensuring that their outcomes and benefits are equitable---regardless of race, gender, age, or socioeconomic status---is imperative.  In particular, AI and ML tools have already become pivotal in the healthcare domain, where public and private healthcare systems increasingly suffer from deficits in healthcare personnel and soaring expenditures and waste \cite{karpathakis2024, patel2023}. Such tools have been one avenue for relieving these burdens by aiding healthcare personnel in completing administrative and clinical tasks (e.g., diagnosis) \cite{aung2021, morley2023}. As a result, the global enthusiasm and investment in health AI has skyrocketed. As of 2024, the global market size of health AI was estimated at \$26.57 billion USD; this number is expected to reach \$187.69 billion USD by 2030 \cite{GrandViewResearch2025}. 

Despite its promise and garnered enthusiasm, AI tools are not a silver bullet and have raised some ethical concerns. Namely, a number of AI tools have been shown to exacerbate the biases embedded in the training data \cite{obermeyer2019, daneshjou2022, park2021, celi2022}. AI models are designed to learn general patterns in human-supplied data. As such, social norms, biases, and errors in this data are learned these tools. These limitations are detrimental in high-stakes fields such as healthcare, where systematic model errors can exacerbate public health challenges (e.g., health disparities), delay intervention, or even deny life-saving care. For example, predictive models of health outcomes often rely on proxies of a patient's true state such as International Classification of Diseases (ICD) codes, treatment records, or healthcare spending. However, these proxies often reflect systematic inequities in access to care \cite{obermeyer2019}. Patients who are never screened or lack access to treatment are often overlooked by these models, thereby perpetuating existing disparities. For instance, even bias-mitigated models ~\cite{park2021} concluded that non-Hispanic white patients are at higher risk for postpartum depression (PPD) than ethnic and racial minorities, a finding that likely reflects unequal screening and treatment rather than lower true prevalence ~\cite{kozhimannil2011, bodnar2017}. These model decisions are due to the fact that expecting mothers belonging to racial and ethnic minority groups are less likely to receive screening and treatment, and therefore are underrepresented in the data \cite{kozhimannil2011, haight2024, dennis2006}. While the current paper is motivated by examples from healthcare, these issues persist in other domains. Large language models (LLMs) used for resume screening have been shown to favor applicants with stereotypically male names over those with stereotypically female names \cite{wilson2024gender}. Mortgage-lending algorithms also tend to approve fewer applications from minority applicants---even in ostensibly race-blind automated underwriting systems \cite{bhutta2021}. 

To address concerns of disproportionate harm, researchers have developed numerous bias-mitigating interventions to encourage algorithmic fairness \cite{friedler2019}. 
Among these is reweighting, a widely used pre-processing technique known for its simplicity and demonstrated effectiveness. Specifically, the current study will assess reweighting as described in Kamiran \& Calders \cite{kamiran2012}. This approach assigns different weights to data points during model training to encourage discrimination-free model behavior without requiring explicit relabeling of records in the dataset. Applying this technique has been shown to improve algorithmic fairness, as defined by improved demographic parity (defined below) and more comparable false negative rates between groups. On the other hand, prior work has also shown that there is a general trade-off between fairness and accuracy \cite{sun2022trade,dutta2020,kim2020,pinzon2022,li2023}. It is well established that enforcing demographic parity when base rates differ between groups raises error rates for some individuals~\cite{hardt2016, corbettdavies2017}. Less attention has been paid to the fact that reweighting, a pre-processing step often applied to reduce disparities in error rates, implicitly enforces such parity, and to the question of which groups bear the resulting cost.

In this paper, we argue that reweighting can have unintended and potentially harmful effects, particularly for groups with a higher baseline prevalence of an event. Postpartum depression illustrates why the direction of this effect depends on the data. If a minority group's true prevalence is higher and its cases are accurately recorded, reweighting will lower that group's predicted risk and increase missed cases, the harm we study here. If instead its cases are under-recorded because its members are less likely to be screened or treated~\cite{haight2024, kozhimannil2011}, the group appears to have \emph{lower} prevalence in the data, and reweighting may partially correct that bias. Our simulations use accurate labels and therefore isolate the first case. Through a series of carefully controlled simulations, we demonstrate how dataset characteristics shape the impact of reweighting on model performance and fairness outcomes. Our findings offer critical insights into the limitations of reweighting and underscore the importance of considering dataset structure and prediction goals when implementing fairness interventions. In particular, reweighting may yield misleading improvements in fairness (e.g., improved demographic parity, reduced differences in false negative rates between groups) by underestimating outcomes for one group while improving for another. By unpacking when and how reweighting may fail, we aim to equip practitioners and researchers with a more nuanced understanding of its risk, helping them make informed decisions in the pursuit of equitable AI.

\subsection{Assessing Algorithmic Fairness}
Numerous metrics exist to quantify the fairness of model predictions \cite{Weerts_Fairlearn_Assessing_and_2023}, a selection of which are discussed in this section. Fairness is often conceptualized at two levels: group and individual. Group fairness seeks to ensure that performance metrics—such as the likelihood of a positive recommendation or prediction accuracy—are comparable across demographic groups (e.g., gender, race, or ethnicity). Individual fairness aims to ensure that individuals with similar characteristics receive similar recommendations. It is important to note that the choice of metric is highly context-sensitive and should be guided by the specific goals of the application. The metrics evaluated in the present study are motivated by healthcare applications, where appropriate identification or recommendation are critical, and where false negatives may be life-threatening. In other domains, such as criminal justice, algorithmic fairness may instead prioritize minimizing false positives between groups (e.g., the COMPAS algorithm was more likely to incorrectly predict re-offense for Black individuals than for White individuals \cite{Angwin_Larson_Mattu_Kirchner_2016}).

For the purposes of illustrating the effects of reweighting in healthcare and similar domains, we focus on two group-level fairness metrics that this technique directly influences: Demographic Parity (DP) and False Negative Rate Parity (FNR Parity). Let $\hat{Y}\in \{0,1\}$ indicate the predicted class, $Y\in \{0,1\}$ indicate the true/observed class, and $G\in \{A, B\}$ indicate group membership--also known as sensitive attributes or features such as race and sex. 



Demographic Parity (DP) aims to ensure that a model's predictions are independent of a sensitive attribute (e.g., race or gender). In other words, the probability of a positive prediction should be equal across different demographic groups. In classification problems, DP can be expressed as:
\begin{align*}
    DP \coloneqq P(\hat{Y}=1 | G=A) - P(\hat{Y}=1 | G=B) 
\end{align*}
Values at zero indicate parity (i.e., fairness), while deviations indicate the degree of disparity (i.e., bias); values closer to zero indicate better fairness. However, DP is not always appropriate, such as when there are true variations in event prevalence between groups. For example, the incidence rate of breast cancer is higher in females than in males due to hormonal differences between the groups \cite{anderson2010breastcancer}. As such, other fairness metrics emphasize parity in model performance between groups. One such metric is False Negative Rate (FNR) Parity which aims to ensure equal miss rates across groups. Again, the choice of metric is context dependent. However, in many healthcare applications, which is the motivating example behind these simulations, FNR parity is crucial to avoid perpetuating many of the existing health disparities---health disparities typically involve certain groups missing out on care. It can be expressed as:
\begin{equation*}
    \text{FNR Parity} \coloneqq P(\hat{Y}=0|Y = 1, G = A) - P(\hat{Y}=0|Y = 1, G = B) 
\end{equation*}



\subsection{Methods for improving algorithmic fairness}
Reweighting is a model-agnostic pre-processing technique used to improve likelihood of identification or recommendation for groups with a lower baseline likelihood of event without requiring re-labeling of the training data. As described in Kamiran \& Calders \cite{kamiran2012}:
\begin{quote}
  Instead of changing the labels, the tuples in the training dataset are assigned weights. As we will show, by carefully choosing the weights, the training dataset can be made discrimination-free w.r.t. \textit{S} [sensitive attribute] without changing any labels. The weights on the tuples can be used directly in any method based on frequency counts.
\end{quote}

Assuming independent, categorical sensitive attributes ($S$) and classes ($C$), the weight for each data point, \textit{i}, in the dataset is calculated as:

\begin{equation*}
    w_i \coloneqq \frac{P_{exp}(S = s_i \; \cap \; C = c_i)}{P_{obs}(S = s_i \; \cap \; C = c_i)} = \frac{P_{obs}(S = s_i) \cdot P_{obs}(C = c_i)}{P_{obs}(S = s_i \; \cap \; C = c_i)},
\end{equation*}

\

\noindent where $s_i$ and $c_i$ are the values of the sensitive attribute and the true target value, respectively, for the $i-$th data point. The numerator in the expression is the expected probability of observing an individual belonging to the group ($s_i$) and being in the class ($c_i$), assuming independence between sensitive attribute ($S$) and class ($C$). The denominator is the observed probability in the dataset. If the observed probability is lower than the expected probability under assumptions of independence, then that individual will be assigned a weight greater than 1. If the observed probability is greater than the expected probability under assumptions of independence, then that individual will be assigned a weight below 1. If the observed and expected probabilities are equal, then that individual will be assigned a weight of 1. These weights are then used to determine the relative contribution of each data point to the model's learning. As such, these weights are expected to impact classification across groups (based on sensitive attributes), as well as performance.

\subsection{Results of reweighting in real-data applications}
Numerous real-data applications of reweighting have demonstrated its ability to ameliorate underestimation---specifically by increasing the probability of positive predictions for groups with lower baseline probabilities \cite{park2021,wong2024,li2022,nyawambi2024,dalvi2024}. Similarly, reweighting has also been shown to improve false negative rates for groups in which models initially under-estimated risk \cite{wong2024}. Other work has also found that reweighting can improve balanced accuracy of Decision Tree (DT) and reduces bias across various metrics (e.g., Average Odds Difference (AOD), Equal Opportunity Difference (EOD), and Theil Index (TI)), but does not improve fairness of K Nearest Neighbor (KNN) \cite{blow2024}. However, determining the conditions under which reweighting is effective remains a challenge in real-data applications given that the data are not free from societal biases.

\subsection{Known Trade-offs}
A commonly-discussed limitation of bias mitigation techniques, such as reweighting, is the trade-off between fairness and overall model performance (e.g., \cite{sun2022trade,dutta2020,kim2020,pinzon2022,li2023}). In many cases, enhancing fairness metrics can lead to reduced predictive accuracy or other performance measures. However, a decline in accuracy is not necessarily problematic when the ground-truth labels reflect societal inequities or historical bias. In such cases, there should be no aspiration to reproduce incorrect or unjust outcomes.

For example, in credit lending, Black applicants have historically been less likely to receive loan approval relative to White applicants despite equivalent qualifications \cite{bhutta2021}. A model trained with reweighted samples may adjust its recommendations to increase approvals for Black applicants. Although these predictions may diverge from the historical “true” labels, this divergence can be desirable—since those labels encode discrimination rather than genuine creditworthiness.

More directly related to our work, enforcing demographic parity when base rates differ is known to raise error rates for some individuals~\cite{hardt2016, corbettdavies2017}. Prior studies have examined how reweighting and related target-balancing methods interact with error-rate metrics~\cite{kehrenberg2020}, how multiple fairness metrics can be jointly balanced~\cite{hsu2022}, and how data and model biases interact to produce fairness--fairness trade-offs~\cite{pombal2022}. These studies establish that such trade-offs exist. Our focus is on how dataset characteristics, specifically group size and baseline prevalence, determine which group bears the cost of a widely used pre-processing method, and by how much.

\subsection{Present Study}
Despite evaluating the benefits of reweighting as a bias mitigation tool, to our knowledge, previous work has yet to systematically examine how effective reweighting is as a function of dataset characteristics such as diversity (presence of majority and minority groups) and baseline prevalence rates. We make three contributions. First, we show that a widely used pre-processing method carries the cost of demographic parity implicitly, even when it is applied to address disparities in error rates. Second, we show that which group bears this cost, and how much, is predictable from group size and baseline prevalence: it falls most heavily on small groups with higher prevalence. Third, we show that this cost is invisible in overall accuracy, so that reweighting can appear to improve fairness at no cost while shifting errors onto the smaller group. Thus, our current study's objective is to demonstrate the benefits and potential harms of reweighting through carefully controlled simulations, in which the synthetic data are free of societal influence, allowing us to study the effect of reweighting across data characteristics in isolation.

\subsection{Hypotheses}
Given DP's aspiration towards model predictions that are independent of a sensitive attribute, reweighting as proposed by Kamiran \& Calders \cite{kamiran2012} should reliably improve DP as has been demonstrated in several papers \cite{park2021,wong2024} (\textit{\textbf{H1}}). Reweighting can also be applied to address disproportionate miss (FNR) and false alarm rates between groups (e.g., Wong et al. \cite{wong2024}, Chen et al. \cite{chen2023}) (\textit{\textbf{H2}}). While reweighting has demonstrably improved positive prediction rates for groups with lower baseline rates of event, the formulation also suggests that it can \textit{decrease} positive prediction rates for groups with higher baseline rates of event (\textit{\textbf{H3a}}) thus increasing false negative rates (\textit{\textbf{H3b}}), which may be an undesirable consequence. We demonstrate these limitations through a carefully controlled simulation study. 

\section{Methods}
All simulations were performed in Python (version 3.12.16) using Numpy (version 2.5.4) and scikit-learn (version 1.9.1). The corresponding Python code is available at: \href{https://anonymous.4open.science/r/reweighting-sims/}{https://anonymous.4open.science/r/reweighting-sims/}.

\textbf{Simulation design.} The sensitive feature in the present simulation study was group membership, which consisted of two possible groups: $A$ and $B$. The six experimental conditions outlined in Table \ref{tab:exp_cond} manipulated two dataset characteristics: dataset diversity and relative event likelihood. Dataset diversity was determined by the proportion of the total sample size made up of Group $A$ (e.g., 0.20); the remaining was made up of Group $B$ (e.g., 0.80). Relative event likelihood was determined by the regression parameter associated with the group predictor, $b_a$; a positive value indicated that Group $A$ was \textit{more} likely than Group $B$ to exhibit the positive outcome, a negative value indicated that Group $A$ was \textit{less} likely, and a value of zero indicated no difference in event likelihood between groups. Table~\ref{tab:sub_ex} lists a substantive example for each condition. To assess the effect of reweighting in each of the six conditions, models were fit to the simulated data in each condition with and without reweighting as described in Kamiran \& Calders \cite{kamiran2012}. 

\begin{table}[h]
\begin{threeparttable}
\caption{Experimental conditions varying two data characteristics: dataset diversity and relative event likelihood.}
    \label{tab:exp_cond}
    \centering
    \begin{tabular}{l l l l l}
    \hline
         \# & Proportion Group A & Relative Event Likelihood in Group A & Prevalence A & Prevalence B \\
         \hline
         1 & 0.2 (Minority) & Less Likely ($b_a = -0.7$) & 7.0\% & 10.0\% \\
         2 & 0.5 (Equal) & Less Likely ($b_a = -0.7$) & 7.0\% & 10.0\%  \\
         3 & 0.2 (Minority) & Equally Likely ($b_a = 0$) & 10.0\% & 10.0\% \\
         4 & 0.5 (Equal) & Equally Likely ($b_a = 0$) & 10.0\% & 10.0\%  \\
         5 & 0.2 (Minority) & More Likely ($b_a = +0.7$) & 13.5\% & 10.0\%  \\
         6 & 0.5 (Equal) & More Likely ($b_a = +0.7$) & 13.5\% & 10.0\% \\
        \hline
    \end{tabular}
    
    \begin{tablenotes}
    \small 
    \item \textit{Note:} Prevalence is the mean realized prevalence across 3,000 repetitions. Each condition drew 10,000 observations, split 70/30 into 7,000 training and 3,000 test observations.
    \end{tablenotes}
\end{threeparttable}
\end{table}

\begin{table}[h]
    \caption{Experimental conditions and their corresponding substantive examples.}
    \label{tab:sub_ex}
    \centering
    \begin{tabular}{l l l l l}
    \hline
         \# & Proportion Group A & Relative Event Likelihood in Group A & Substantive Example \\
         \hline
         1 & 0.2 (Minority) & Less Likely & Mortgage Approval \cite{bhutta2021} \\
         2 & 0.5 (Equal) & Less Likely & Resume Selection \cite{wilson2024gender} \\
         3 & 0.2 (Minority) & Equally Likely & Control Group \\
         4 & 0.5 (Equal) & Equally Likely & Control Group \\
         5 & 0.2 (Minority) & More Likely & Postpartum Depression $^{a}$ \cite{wong2024}  \\
         6 & 0.5 (Equal) & More Likely & Breast Cancer \cite{anderson2010breastcancer} \\
    \hline
    \end{tabular}
    \\[2pt]
    {\small Note: Examples indicate the direction of the difference in event likelihood, not its mechanism or size; for instance, the difference in breast cancer incidence by sex is far larger than the differences simulated here. Mortgage approval and resume selection are cases in which observed outcomes partly reflect discrimination (label bias), which our simulations do not model. $^{a}$Whether true prevalence of postpartum depression differs by race and ethnicity is debated~\cite{liu2017}; because members of racial and ethnic minority groups are less likely to receive screening and treatment~\cite{haight2024, kozhimannil2011}, observed labels may undercount their cases, in which case this example corresponds to Condition~1 rather than Condition~5 (see Section~\ref{sec:discussion}).}
\end{table}

\textbf{Data generation.} For each observation $i$, the binary outcome was generated from the logistic regression model:
\[
Y_i \sim \text{Bernoulli}(p_i), \qquad \ln\!\left(\frac{p_i}{1-p_i}\right) = b_0 + b_a X_{a,i} + b_1 X_{1,i} + b_2 X_{2,i} + b_3 X_{3,i},
\]
where $X_a \in \{0, 1\}$ indicated whether an individual belonged to Group $A$ ($X_a = 1$) or Group $B$ ($X_a = 0$). The model also included three independent covariates: $X_1$ was normally distributed with mean 0 and standard deviation 1, $X_2$ was normally distributed with mean 5 and standard deviation 2, and $X_3$ was drawn from a discrete uniform distribution on 0--20. The coefficients $b_1$, $b_2$, and $b_3$ were $-0.5$, $-0.3$, and $0.8$. 
Because the covariates are not centered at zero, setting the intercept to $\ln(0.10/0.90)$ would give a 10\% probability only to an individual with $X_1 = X_2 = X_3 = 0$; averaged over the covariate distribution, the prevalence would be approximately 74\%. We therefore computed $b_0$ so that the expected prevalence in Group $B$ ($X_a = 0$) was 10\%:
\begin{equation}\label{eq:calibration}
\mathbb{E}_{X_1, X_2, X_3}\!\left[\frac{1}{1 + e^{-(b_0 + b_1 X_1 + b_2 X_2 + b_3 X_3)}}\right] \approx 0.10 .
\end{equation}
We approximated the expectation by averaging over 2,000,000 covariate draws generated with a fixed random seed, and solved Equation~\ref{eq:calibration} for $b_0$ using Brent's root-finding method, giving $b_0 \approx -13.48$. The same value was used in every condition and repetition. Because Group $B$ is defined by $X_a = 0$, $b_a$ does not affect this calibration, and Group $A$'s prevalence follows from its log-odds differing by $b_a$: approximately 7\%, 10\%, or 13.5\% (Table~\ref{tab:exp_cond}).

\textbf{Training and test data.} In each repetition and condition, 10,000 observations were simulated and split into training (70\%) and test (30\%) sets, stratified by group and outcome. Reweighting weights ~\cite{kamiran2012} were computed on the training set only. Covariates were standardized using training-set means and standard deviations. 

\textbf{Classifiers.} Six classifiers, implemented in scikit-learn with default settings unless stated, were fit to the same training data within a repetition, with and without reweighting: unpenalized logistic regression ($C = 10^{10}$), L2-regularized logistic regression (C=1), a decision tree (minimum 20 observations per leaf), a random forest (200 trees, minimum 5 per leaf), histogram gradient boosting, and a multilayer perceptron (hidden layers of 32 and 16 units). Reweighting was applied through each model's sample weights. Predicted probabilities were classified at a threshold of 0.5. 

\textbf{Evaluation and statistical analysis.} For each model and condition, we computed on the held-out test set the positive-prediction rate, false negative rate (FNR), and false positive rate for each group; DP and FNR parity (Group $A$ minus Group $B$); and overall accuracy and AUROC. Additional missed cases per 10,000 group members were computed as the change in FNR multiplied by the group's prevalence and by 10,000. The share of a narrowed FNR gap attributable to the higher-baseline group was its increase in FNR divided by the total reduction in the gap. The simulation was repeated 3,000 times. Each repetition used its own random seed, shared by all conditions, and within each repetition the unweighted and reweighted models were fit to identical data. Effects are therefore reported as the mean paired difference (reweighted minus unweighted) with 95\% confidence intervals and tested with two-sided Wilcoxon signed-rank tests. Comparisons between group sizes were also paired by repetition. Because 3,000 paired repetitions make even negligible differences statistically significant, we interpret effect sizes rather than $p$-values.

\section{Results}
\label{results}

\subsection{Demographic parity (DP)}
\begin{figure}[h]
  \centering
  \includegraphics[width=\linewidth]{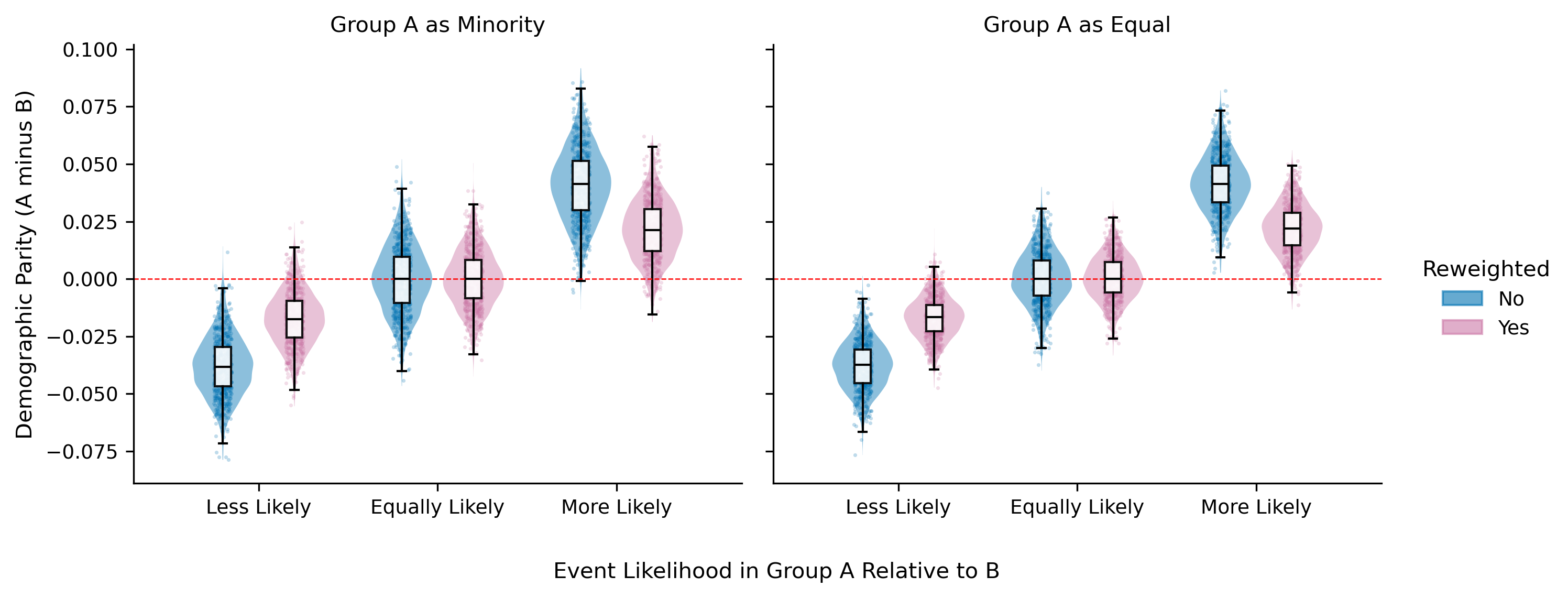}
  \caption{The effect of reweighting on Demographic Parity (DP) (logistic regression; held-out test set).}
  \label{fig:dp}
\end{figure}

\begin{figure}[h]
  \centering
  \includegraphics[width=\linewidth]{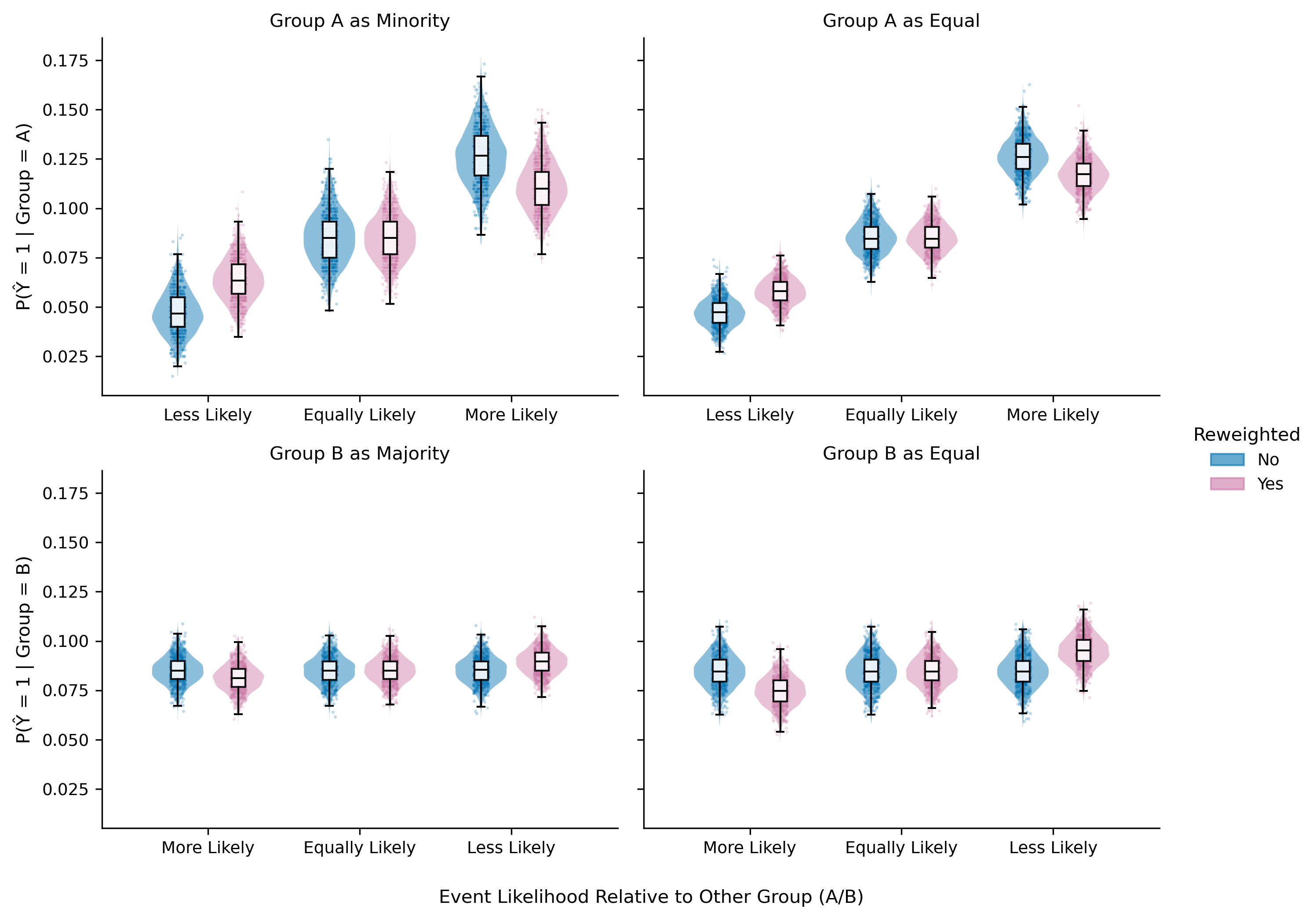}
  \caption{The effect of reweighting on model predictions (logistic regression; held-out test set).}
  \label{fig:pp}
\end{figure}

We replicated the previously observed effect of reweighting on DP as shown in Figure \ref{fig:dp}. The DP values from models without reweighting are shown in blue, and the DP values with reweighting are shown in pink. As expected, reweighting made no significant difference between groups when the event was equally likely between groups. When event prevalence varied between groups, reweighting promoted DP as indicated by the pink distributions being drawn closer to zero (\textit{\textbf{H1}} supported): it roughly halved the difference in positive-prediction rates, from $-3.8$ to $-1.7$ percentage points when Group $A$ was less likely to experience the event, and from $4.1$ to $2.1$ percentage points when it was more likely, although it did not eliminate it. However, it is important to highlight a potentially undesired consequence of reweighting, especially for groups in which the baseline likelihood of the event is more likely. While reweighting increased the probability of a positive prediction for groups with lower baseline likelihoods, it also \textit{decreased} the probability of a positive prediction for groups with higher baseline likelihoods. Figure \ref{fig:pp} provides a more detailed breakdown of the proportion of positive predictions for Groups $A$ and $B$ across each of the six experimental conditions. Table \ref{tab:pp_results} shows the change in proportion of positive prediction before and after reweighting in groups with higher baselines (\textit{\textbf{H3a}} supported). For a minority group with the higher baseline, the positive-prediction rate fell from 12.6\% to 11.1\% ($-1.5$ percentage points, a 12\% relative decrease). The same pattern held across all six classifiers (Section~\ref{sec:models}). 

The decrease in probability of positive prediction is also significantly smaller for the majority group ($-0.4$ percentage points) compared to the minority ($-1.5$; $p < .001$) and equal-sized groups ($-0.9$ to $-1.0$; $p's < .001 $). The decrease was likewise significantly smaller for equal-sized groups than for the minority group ($p's < .001$). These comparisons were paired by repetition, since conditions within a repetition shared the same random seed. In summary, the unintended and undesirable consequences of reweighting were most pronounced in smaller, non-majority groups with higher baseline event prevalence.

\begin{table}[h]
\caption{The effect of reweighting on the probability of positive classification by group size and relative event likelihood (logistic regression, held-out test set). Before and after are mean rates; $\Delta$ is the mean paired difference (reweighted minus unweighted) across 3,000 repetitions, with 95\% confidence intervals (CI) and two-sided \textit{p}-values from Wilcoxon signed-rank tests. Values correspond to Figure \ref{fig:pp}.}
\label{tab:pp_results}
    \centering
    \small
    \setlength{\tabcolsep}{5pt}
    \begin{tabular}{l l l c c c c c}
        \hline
        Group & Group & Relative Event & & & & & Two-sided \\
        Label & Size  & Likelihood     & Before & After & $\Delta P(\hat{Y}=1|\cdot)$ & 95\% CI & \textit{p}-value \\
        \hline
         A & Minority & More Likely & 0.1263 & 0.1109 & $-0.0154$ & [$-0.0156$, $-0.0152$] & $<.001$ \\
         A & Equal    & More Likely & 0.1264 & 0.1172 & $-0.0092$ & [$-0.0094$, $-0.0091$] & $<.001$ \\
         B & Equal    & More Likely & 0.0849 & 0.0748 & $-0.0101$ & [$-0.0102$, $-0.0100$] & $<.001$ \\
         B & Majority & More Likely & 0.0853 & 0.0813 & $-0.0040$ & [$-0.0040$, $-0.0039$] & $<.001$ \\
        \hline
         A & Minority & Less Likely & 0.0472 & 0.0640 & $+0.0167$ & [$+0.0165$, $+0.0170$] & $<.001$ \\
         A & Equal    & Less Likely & 0.0472 & 0.0580 & $+0.0108$ & [$+0.0107$, $+0.0109$] & $<.001$ \\
         B & Equal    & Less Likely & 0.0849 & 0.0952 & $+0.0104$ & [$+0.0103$, $+0.0105$] & $<.001$ \\
         B & Majority & Less Likely & 0.0852 & 0.0895 & $+0.0043$ & [$+0.0042$, $+0.0044$] & $<.001$ \\
        \hline
    \end{tabular}
\end{table}

\subsection{False negative rate (FNR)}
\begin{figure}[h]
  \centering
  \includegraphics[width=\linewidth]{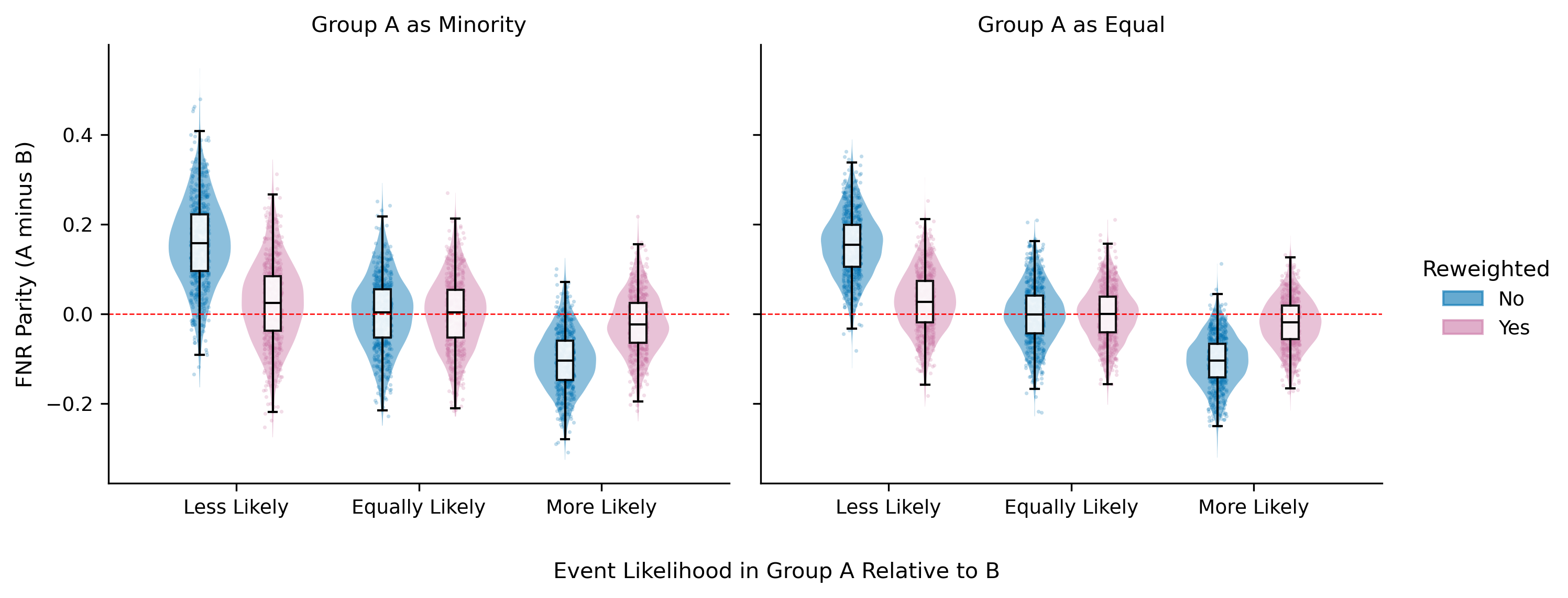}
  \caption{The effect of reweighting on false negative rates (FNR) parity (logistic regression; held-out test set.}
  \label{fig:fnr_parity}
\end{figure}

\begin{figure}[h]
  \centering
  \includegraphics[width=\linewidth]{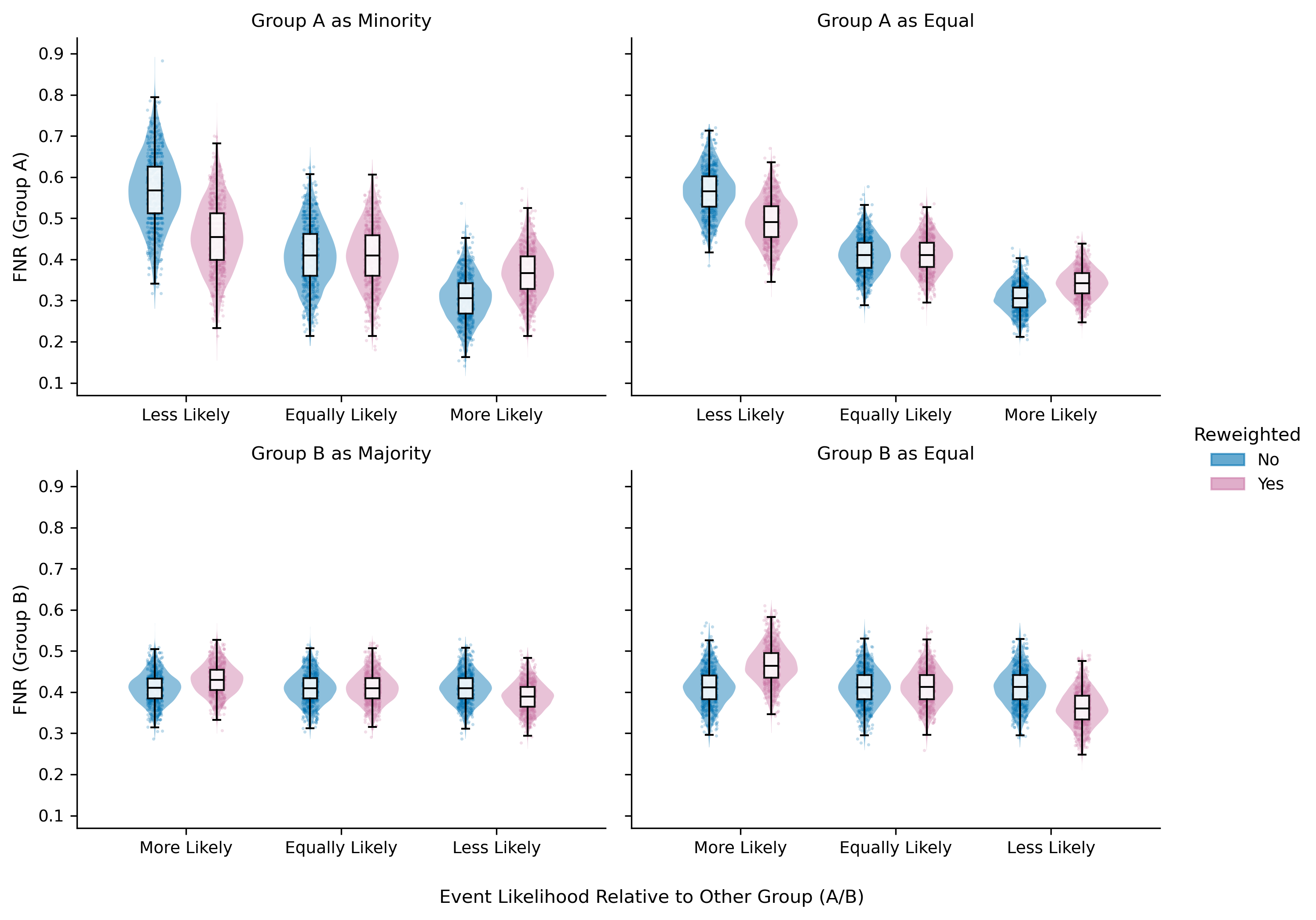}
  \caption{The effect of reweighting on false negative rates (FNR) (logistic regression, held-out test set).}
  \label{fig:fnr}
\end{figure}

The effect of reweighting on FNR and FNR parity is consistent with DP, as illustrated in Figures \ref{fig:fnr_parity} and \ref{fig:fnr}. Like for DP, values from models without reweighting are shown in blue, and values with reweighting are shown in pink. As expected, reweighting made a negligible difference when the event was equally likely between groups: group FNRs changed by 0.1 percentage points or less, and FNR parity did not change ($p\text{'s} > .05$). 
When event prevalence varied between groups, reweighting promoted FNR parity, as indicated by the pink distributions being drawn closer to zero (see Figure \ref{fig:fnr_parity}): the difference in FNR between groups fell from 15.8 to 2.6 percentage points when group $A$ was less likely to experience the event, and from $-10.3$ to $-2.1$ percentage points when it was more likely (\textit{\textbf{H2}} supported). Reweighting improved (i.e., lowered) the FNR for groups with lower baseline likelihoods of the event, most strongly for minority groups ($-11.2$ percentage points). However, reweighting increased FNRs for groups with higher baseline likelihoods. 
Table \ref{tab:fnr_results} shows the change in FNR before and after reweighting in groups with higher baselines (\textit{\textbf{H3b}} supported). For a minority group with the higher baseline, FNR rose from 30.7\% to 36.8\% ($+6.1$ percentage points), corresponding to approximately 82 additional missed cases per 10,000 members of the group. Across all six classifiers, the increase ranged from 3.9 to 6.2 percentage points (Section~\ref{sec:models}).
Like for DP, the increase in FNR is significantly smaller for the majority group ($+2.0$ percentage points) compared to the minority group ($+6.1$; $p < .001$) and equal-sized groups ($+3.6$ to $+5.3$; $p's < .001$), and significantly smaller for equal-sized groups than for the minority group ($p's < .001$). Among equal-sized groups, the increase was larger for Group $B$ ($+5.3$ percentage points) than for Group $A$ ($+3.6$), even though the difference in log-odds between groups was the same. The increase also depended on how many of a group's positive cases lay near the decision threshold: Group $B$, whose baseline FNR was higher, lost more true positives for a similar shift in predicted risk. Like for DP, the unintended and undesirable consequences of reweighting on FNR were most pronounced in smaller, non-majority groups with higher baseline event prevalence.
When the higher-baseline group was the minority, approximately three-quarters (74\%) of the reduction in the FNR gap between groups came from that group's FNR rising rather than from the other group's falling (42\% for equal-sized groups; 15\% when the higher-baseline group was the majority).

\begin{table}[h]
\caption{The effect of reweighting on FNR for groups with higher baseline likelihoods across group sizes (logistic regression, held-out test set). Values are mean paired differences (reweighted minus unweighted) across 3,000 repetitions, with 95\% confidence intervals (CI) and two-sided \textit{p}-values from Wilcoxon signed-rank tests. Additional missed cases are $\Delta$FNR $\times$ group prevalence $\times$ 10,000, using training-set prevalence. Values correspond to Figure \ref{fig:fnr}.}
\label{tab:fnr_results}
    \centering
    \small
    \setlength{\tabcolsep}{5pt}
    \begin{tabular}{l l l c c c c}
        \hline
        Group & Group & Relative Event & & & Two-sided & Additional missed \\
        Label & Size  & Likelihood     & $\Delta$ FNR & 95\% CI & \textit{p}-value & cases per 10,000 \\
        \hline
         A & Minority & More Likely & $+0.0608$ & [$0.0597$, $0.0618$] & $<.001$ & 82 \\
         A & Equal    & More Likely & $+0.0356$ & [$0.0351$, $0.0362$] & $<.001$ & 48 \\
         B & Equal    & More Likely & $+0.0531$ & [$0.0524$, $0.0539$] & $<.001$ & 53 \\
         B & Majority & More Likely & $+0.0202$ & [$0.0198$, $0.0206$] & $<.001$ & 20 \\
        \hline
    \end{tabular}
\end{table}

\begin{figure}[h]
    \centering
    \includegraphics[width=\linewidth]{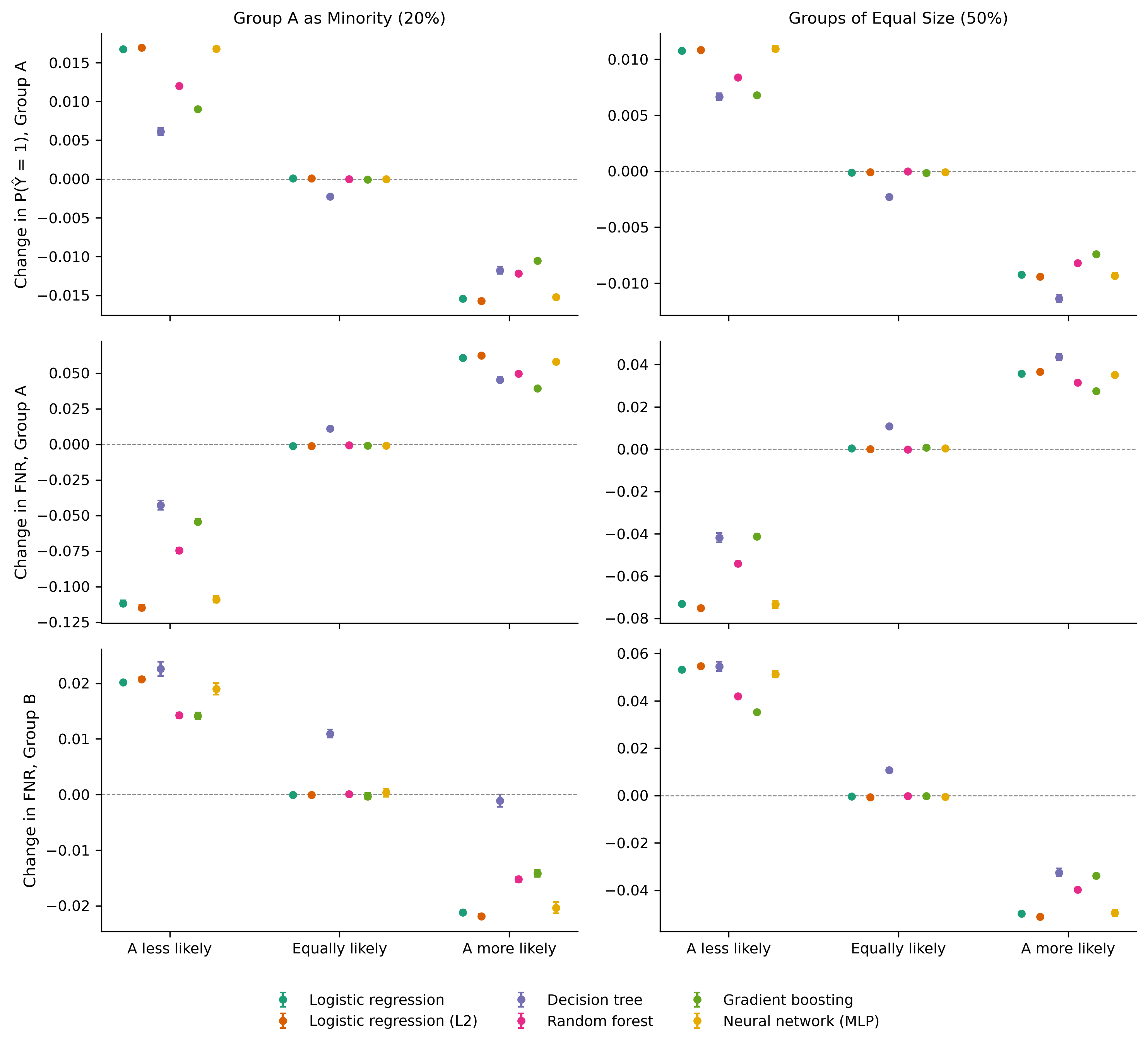}
    \caption{Change after reweighting (reweighted minus unweighted) in the positive-prediction rate of Group $A$ (top), the FNR of Group $A$ (middle), and the FNR of Group $B$ (bottom), for six classifiers. Left: Group $A$ is the minority (20\%); right: groups are of equal size. Points are means across 3,000 repetitions on the held-out test set; error bars are 95\% confidence intervals.}
    \label{fig:models}
\end{figure}

\subsection{Robustness across classifiers}
\label{sec:models}
To test whether these effects are specific to logistic regression, we repeated every analysis with five additional classifiers fit to the same data: L2-regularized logistic regression, a decision tree, a random forest, gradient boosting, and a multilayer perceptron. Figure~\ref{fig:models} shows the change in positive-prediction rate and FNR after reweighting for every classifier and condition, and Table~\ref{tab:models} summarizes the key comparisons.
The pattern held for all six classifiers. Reweighting lowered the FNR of the group with the lower baseline and raised the FNR of the group with the higher baseline in every condition where prevalence differed, with one exception: for the decision tree, the FNR of a lower-baseline majority group did not change ($-0.1$ percentage points, 95\% CI [$-0.2$, $0.0$]). All other changes had 95\% confidence intervals excluding zero. For a higher-baseline minority group, FNR rose by 3.9 to 6.2 percentage points across classifiers, and for a higher-baseline majority group, by 1.4 to 2.3 percentage points. The increase was therefore larger for a minority than for a majority group under every classifier.
The size of the effects differed by classifier. They were largest for the two logistic regressions and the multilayer perceptron, and smaller for the tree-based models, partly because the tree-based models' unweighted FNRs already differed less between groups. For the decision tree, the effect of group size was also weaker: the FNR increase for a higher-baseline minority group ($+4.5$ percentage points) was close to that for an equal-sized group ($+4.4$). Notably, the decision tree showed almost no FNR gap between groups before reweighting when Group $A$ was the minority with the higher baseline ($-0.3$ percentage points), but reweighting created a gap of $4.3$ percentage points in the opposite direction. In this case, reweighting worsened FNR parity rather than improving it.
For every classifier and condition, reweighting changed overall accuracy by less than 0.1 percentage points. The redistribution of errors between groups was therefore invisible in aggregate performance. Tree-based models fit their training data considerably more closely than held-out test data, which is why all results are reported on the held-out test set.

\begin{table}[h]
\caption{Change in FNR after reweighting by classifier (held-out test set). Values are mean paired differences (reweighted minus unweighted) across 3,000 repetitions, with 95\% confidence intervals in brackets; all $p < .001$ (Wilcoxon signed-rank). Higher-baseline groups: Group $A$ as minority (condition 5) and as equal-sized group (condition 6), and Group $B$ as majority (condition 1). Lower-baseline group: Group $A$ as minority (condition 1).}
\label{tab:models}
    \centering
    \small
    \setlength{\tabcolsep}{5pt}
    \begin{tabular}{l c c c c c}
        \hline
        & Test & \multicolumn{3}{c}{Higher-baseline group} & Lower-baseline group \\
        Classifier & AUROC & Minority & Equal & Majority & Minority \\
        \hline
        Logistic regression      & 0.960 & $+0.0608$ & $+0.0356$ & $+0.0202$ & $-0.1117$ \\
                                 &       & {\footnotesize [0.0597, 0.0618]} & {\footnotesize [0.0351, 0.0362]} & {\footnotesize [0.0198, 0.0206]} & {\footnotesize [$-0.1138$, $-0.1097$]} \\[2pt]
        Logistic regression (L2) & 0.960 & $+0.0624$ & $+0.0365$ & $+0.0208$ & $-0.1147$ \\
                                 &       & {\footnotesize [0.0613, 0.0635]} & {\footnotesize [0.0360, 0.0371]} & {\footnotesize [0.0204, 0.0211]} & {\footnotesize [$-0.1168$, $-0.1126$]} \\[2pt]
        Multilayer perceptron    & 0.959 & $+0.0581$ & $+0.0352$ & $+0.0190$ & $-0.1090$ \\
                                 &       & {\footnotesize [0.0567, 0.0594]} & {\footnotesize [0.0343, 0.0362]} & {\footnotesize [0.0180, 0.0201]} & {\footnotesize [$-0.1114$, $-0.1065$]} \\[2pt]
        Random forest            & 0.952 & $+0.0496$ & $+0.0315$ & $+0.0143$ & $-0.0745$ \\
                                 &       & {\footnotesize [0.0485, 0.0508]} & {\footnotesize [0.0309, 0.0321]} & {\footnotesize [0.0138, 0.0147]} & {\footnotesize [$-0.0765$, $-0.0725$]} \\[2pt]
        Decision tree            & 0.930 & $+0.0452$ & $+0.0436$ & $+0.0226$ & $-0.0428$ \\
                                 &       & {\footnotesize [0.0433, 0.0472]} & {\footnotesize [0.0421, 0.0450]} & {\footnotesize [0.0213, 0.0239]} & {\footnotesize [$-0.0460$, $-0.0396$]} \\[2pt]
        Gradient boosting        & 0.951 & $+0.0392$ & $+0.0274$ & $+0.0141$ & $-0.0544$ \\
                                 &       & {\footnotesize [0.0381, 0.0404]} & {\footnotesize [0.0267, 0.0281]} & {\footnotesize [0.0135, 0.0147]} & {\footnotesize [$-0.0562$, $-0.0525$]} \\
        \hline
    \end{tabular}
\end{table}

\section{Discussion}
\label{sec:discussion}
Reweighting is a widely applied pre-processing technique for promoting discrimination-free model behavior without altering training data labels. In real-world applications, reweighting has increased positive prediction rates and reduced false negatives for groups with lower baseline event prevalence ~\cite{park2021, wong2024, dalvi2024, nyawambi2024}, but it has also been associated with underestimation for other groups \cite{wong2024}. Because real data are shaped by societal biases, these studies could not determine when such harms arise. In our paper, we reproduced both patterns in synthetic data free from societal biases, for a correctly specified logistic regression and for five classifiers that were not correctly specified. Reweighting consistently improved parity between groups while worsening outcomes for the group with the higher baseline event prevalence, and this cost was greatest when that group was a minority: its FNR rose by 6.1 percentage points, 82 additional missed cases per 10,000 group members, compared with 2.0 percentage points when it was the majority. These results underscore the pivotal role of data characteristics, specifically group size and baseline prevalence, in shaping the effects of fairness interventions. 

\subsection{Why smaller groups bear more of the cost}
These patterns follow from how reweighting works. Kamiran and Calders' weights make group membership and outcome statistically independent in the weighted training data \cite{kamiran2012}. For a logistic regression that includes group membership, this is equivalent to shifting each group's predicted log-odds by 
\[
\operatorname{logit} P(Y=1) - \operatorname{logit} P(Y=1 \mid G=g),
\]
that is, toward the overall prevalence: groups with above-average prevalence are shifted down and groups with below-average prevalence are shifted up. Because the overall prevalence is dominated by the larger group, it lies closest to that group's prevalence and furthest from the smaller group's, so the smaller group receives the largest shift. In our conditions, the shift for the higher-baseline group was $-0.26$ when it was the minority, $-0.16$ to $-0.18$ when groups were of equal size, and $-0.07$ when it was the majority, matching the pattern of the FNR increases. This identity holds in large samples for a correctly specified, unpenalized logistic regression. Across classifiers, the cost to a higher-baseline minority group exceeded that to a majority group in every case, and the full ordering held for all classifiers except the decision tree.
This also clarifies how our findings relate to known trade-offs in algorithmic fairness. By making group membership independent of the outcome in the training data, reweighting acts as a demographic-parity intervention, and when true prevalence differs between groups, enforcing parity in positive predictions is known to raise error rates for some individuals \cite{hardt2016, corbettdavies2017}. Our contribution is to show that a widely used pre-processing step carries this cost implicitly, even when it is applied to address disparities in error rates rather than to enforce demographic parity, to quantify it, and to show which groups bear it.

\subsection{Implications for practice}
Applying reweighting without accounting for dataset structure, particularly group size and baseline event prevalence, can yield misleading improvements in fairness by underestimating outcomes for one group while enhancing outcomes for another. Practitioners should report group-level error rates alongside parity metrics, and inspect group sizes and prevalence before applying reweighting. Whether a shift in errors is harmful also depends on whether observed labels reflect true need: if a group's labels undercount its true cases, as when members of racial and ethnic minority groups are less likely to be screened for postpartum depression, raising its predicted risk may correct rather than cause harm. When the goal is to equalize error rates, methods that target those rates directly, such as group-specific decision thresholds that equalize FNR \cite{hardt2016}, may avoid forcing equal positive-prediction rates. Evaluating such alternatives under the conditions studied here is an important next step. More broadly, there is a pressing need for bias mitigation methods that explicitly account for group-level heterogeneity and are robust across diverse data conditions.

\subsection{Limitations}
Our simulations were designed to isolate the mechanism by which reweighting redistributes errors, and this design has limits. We studied a binary outcome, a single binary sensitive attribute, and one reweighting method \cite{kamiran2012}. Outcomes were generated from a logistic model with independent covariates. Because our covariates were independent of group membership, the models could distinguish groups only through the group indicator itself. In clinical practice, the use of race as a model input is contested and often removed~\cite{vyas2020}, in which case group differences are carried by correlated proxies; the effect of reweighting may then differ, since theoretical work suggests that fairness interventions affect groups less predictably when the sensitive attribute is not used in decisions~\cite{yang2026}.  Although five of our six classifiers did not match this process, real data involve correlated predictors, nonlinear relationships, missing data, and label noise, none of which we modeled. We varied group size and prevalence over a limited set of values, used a single classification threshold of 0.5, and did not examine intersectional groups, where small groups and extreme weights may make the effects larger. We did not compare reweighting with alternative mitigation strategies or apply it to real-world datasets. Additionally, we did not optimize any model parameters for a given dataset. Finally, whether a given shift in errors constitutes harm depends on the application and on whether observed labels reflect true need. Future work should extend these analyses to real-world data, multiple and intersectional sensitive attributes, alternative thresholds, and other mitigation methods.

\section{Key Takeaways}
Reweighting is a convenient and accessible technique to mitigate algorithmic bias. It may be an effective technique when 1) the likelihood of recommendation (i.e., positive prediction) should be equal between groups, 2) there is evidence of bias that leads to over-recommendation in one group, \textit{and} 3) under-recommendation in another group. In other conditions, reweighting may introduce undesirable artifacts. In particular, reweighting may not be an effective technique for increasing recommendations for one group \textit{without} decreasing recommendation for another, and its costs fall most heavily on small groups with higher baseline prevalence. Because these costs are invisible in parity metrics and overall accuracy, practitioners should examine group-level error rates before and after applying it. Our simulations reproduced consequences of reweighting previously observed in real-data applications and provide a controlled setting for understanding its potential pitfalls for fairness. 

\clearpage
\section*{Generative AI Usage Statement}
The authors used Claude (Anthropic; Claude Opus 5.5) for this manuscript. It was used to (1) review the simulation code and identify errors in the original implementation; (2) revise simulation and analysis code, including the held-out evaluation, the additional classifiers, and the scripts that produce the figures and tables; (3) edit text, including the abstract and parts of the Methods, Results, and Discussion, and format tables in LaTeX. The authors directed all analyses, ran the simulations, reviewed and revised all code and text, and verified the reported results and references. The authors take full responsibility for the content of this paper.

\section*{Ethical Considerations Statement}
This study used only simulated data; it involved no human participants, personal data, or deployed systems, and the groups studied are abstract rather than representations of real populations. The main potential adverse impact concerns how the findings are used. Our results show that reweighting can shift errors onto smaller groups with higher baseline prevalence, but they should not be read as an argument against fairness interventions in general, nor as estimates of harm in any particular real-world application, where the size and direction of effects will depend on the data. In particular, when observed labels undercount a group's true cases, as can occur when some groups are less likely to be screened or diagnosed, reweighting may correct rather than cause harm. To support appropriate use, we report effects alongside the conditions that produce them, recommend that practitioners examine group-level error rates rather than parity metrics alone, and release our code so that others can test these effects under their own data conditions.

\clearpage
\bibliographystyle{ACM-Reference-Format}
\bibliography{references}


\end{document}